\documentclass[twoside,11pt]{article}

\usepackage[accepted]{melba}

\usepackage{mwe} % to get dummy images, only for the example
\usepackage{amsmath,amsfonts}

\usepackage{upgreek}
\usepackage{multirow}

\newcommand{\TP}{\mathrm{TP}}
\newcommand{\FP}{\mathrm{FP}}
\newcommand{\FN}{\mathrm{FN}}
\newcommand{\IoU}{\mathrm{IoU}}
\melbaheading{2022:015}{https://www.melba-journal.org/papers/2022:015.html}{2021}{1-24}{09/2021}{07/2022}{Arnavaz, Krause, Zepf, Bærentzen, Krivokapic, Heilmann, Nyeng and Feragen}{}{}

\ShortHeadings{Quantifying Topology In Pancreatic Tubular Networks}{Arnavaz et al}
\firstpageno{1}

\title{Quantifying Topology In Pancreatic Tubular Networks From Live Imaging 3D Microscopy}

\author{\name Kasra Arnavaz \email kasra@di.ku.dk \\  % start right after \author{, or there will be an extra space
	\addr Department of Computer Science, University of Copenhagen, Copenhagen, Denmark
	\AND
	\name Oswin Krause \email oswin.krause@di.ku.dk \\
	\addr Department of Computer Science, University of Copenhagen, Copenhagen, Denmark
	\AND
	\name Kilian Zepf \email kmze@dtu.dk\\
	\addr DTU Compute, Tecnhical University of Denmark, Lyngby, Denmark
	\AND
	\name Jakob Andreas Bærentzen \email janba@dtu.dk\\
	\addr DTU Compute, Tecnhical University of Denmark, Lyngby, Denmark
	\AND 
	\name Jelena M. Krivokapic \email jelena.krivokapic@sund.ku.dk\\
	\addr DanStem, University of Copenhagen, Copenhagen, Denmark
	\AND
	\name Silja Heilmann
	\email silja.heilmann@sund.ku.dk\\
	\addr DanStem, University of Copenhagen, Copenhagen, Denmark
	\AND
	\name Pia Nyeng*
	\email pnyeng@ruc.dk\\
	\addr Department of Science and Environment, Roskilde University College, Roskilde, Denmark
	\AND
	\name Aasa Feragen*
	\email afhar@dtu.dk\\
	\addr DTU Compute, Tecnhical University of Denmark, Lyngby, Denmark\\
	\noindent * Shared last-authorship
}

\begin{document}

% top matter
\maketitle

% abstract
\begin{abstract}%   <- trailing '%' for backward compatibility of .sty file
Motivated by the challenging segmentation task of pancreatic tubular networks, this paper tackles two commonly encountered problems in biomedical imaging: Topological consistency of the segmentation, and expensive or difficult annotation. Our contributions are the following: a)  We propose a topological score which measures both topological and geometric consistency between the predicted and ground truth segmentations, applied to model selection and validation. b) We provide a full deep-learning methodology for this difficult noisy task on time-series image data. In our method, we first use a semisupervised U-net architecture, applicable to generic segmentation tasks, which jointly trains an autoencoder and a segmentation network. We then use tracking of loops over time to further improve the predicted topology. This semi-supervised approach allows us to utilize unannotated data to learn feature representations that generalize to test data with high variability, in spite of our annotated training data having very limited variation. Our contributions are validated on a challenging segmentation task, locating tubular structures in the fetal pancreas from noisy live imaging confocal microscopy. We show that our semi-supervised model outperforms not only fully supervised and pre-trained models but also an approach which takes topological consistency into account during training.  Further, our approach achieves a mean loop score of 0.808 for detecting loops in the fetal pancreas, compared to a U-net trained with clDice with mean loop score 0.762.
\end{abstract}

% keywords
\begin{keywords}
	topological consistency, semisupervised, segmentation, tubular, confocal microscopy
\end{keywords}

\section{Introduction}

Segmentation of tubular structures is a common task in medical imaging, encountered when analyzing structures such as e.g.~blood vessels (\cite{zhang2019attention}), airways (\cite{qin2019airwaynet}), ductal systems (\cite{wang2020deep}), neurons (\cite{li2019precise}). Segmentation is often a first step towards analysing the topological network structure of the organ, and it is therefore crucial that the segmented network structure is reliable. 

\paragraph{Pancreatic tubes}
This paper aims to solve a challenging tubular segmentation task from live fluorescence microscopy imaging. The pancreas produces enzymes and hormones, most importantly insulin which is produced by the so-called $\upbeta$-cells. Since insulin plays a crucial role in regulating blood sugar, malfunction of $\upbeta$-cells could lead to the development of diabetes.

\begin{figure}
\centering
\includegraphics[width=0.6\linewidth]{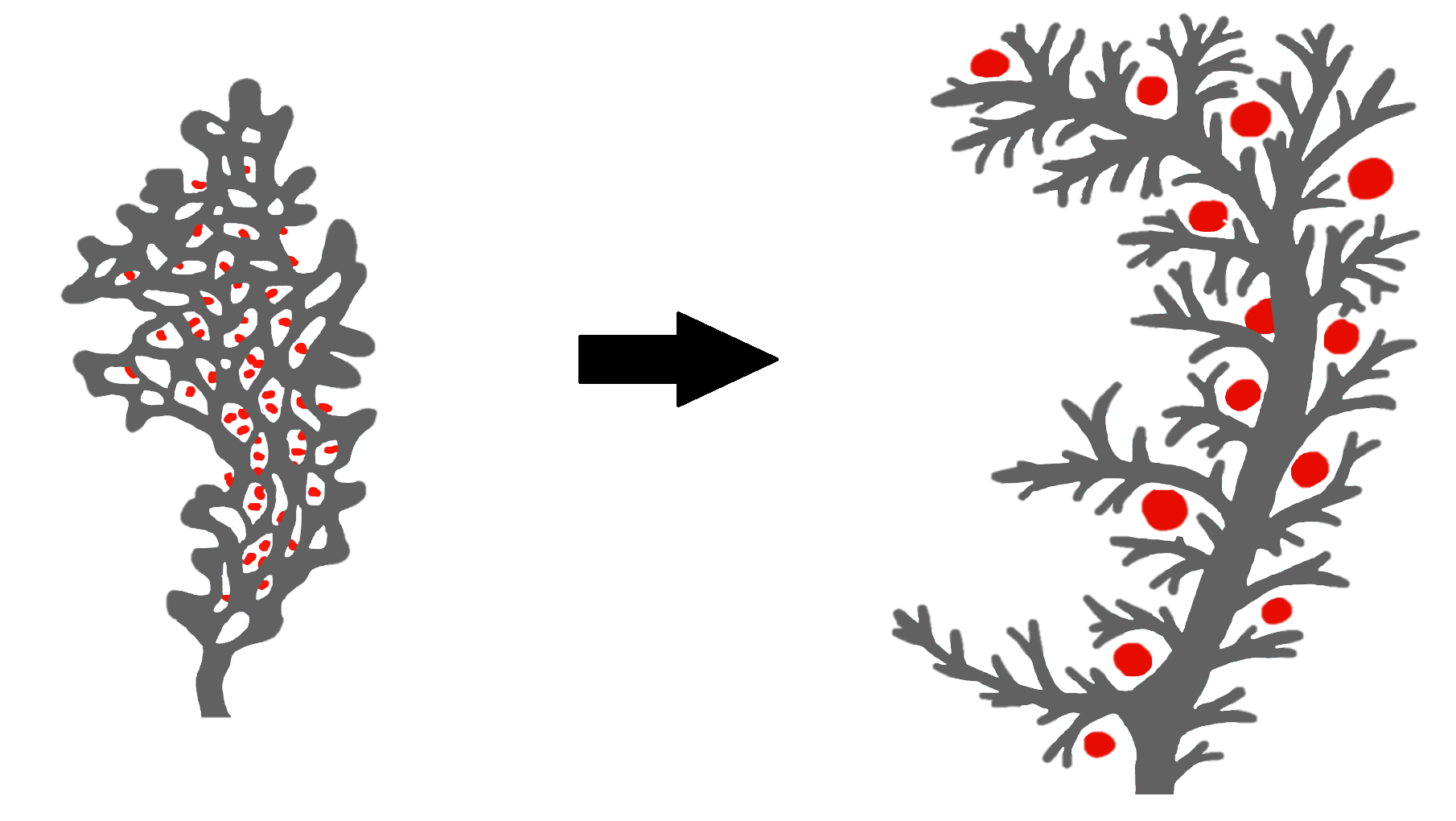}
\caption{Tubular remodeling in pancreas during embryonic development. Gray areas represent tubes, while red dots represent $\upbeta$-cells. }
\label{fig:remodeling}
\end{figure}

The mammalian pancreas contains a tubular ductal network which transports enzymes into the intestines. Unlike superficially similar tubular structures such as the lung airways, the pancreatic tubes do not form by stereotypic branching. Rather, during  embryonic  development,  the pancreatic tubes form by fusion of smaller lumens into a complex web-like network with many loops, which remodel to  a  tree-like structure, as discussed by  ~\cite{pan2011pancreas} and~\cite{villasenor2010epithelial} and illustrated in Figure \ref{fig:remodeling}. As a result, the pancreatic tubes are complex and the overall structure varies from individual to individual, making segmentation and analysis of the network a hard task. 

It has been suggested by \cite{kesavan2009cdc42} and ~\cite{bankaitis2015feedback} that the emergence of $\upbeta$-cells depends on the remodeling of the tubular structure during embryonic development. To quantitatively verify this hypothesis, we require topologically accurate detection of the pancreatic tubes from noisy live imaging confocal 3D microscopy. In particular, we are interested in opening and closure of loops over time as the feature which might affect the emergence of $\upbeta$-cells. We tackle this problem for a dataset consisting of time-lapse 3D images from mouse pancreatic organs, which were recorded as the organs underwent the process of tubular re-organization and emergence of $\upbeta$-cells.

\begin{figure}
\centering
\includegraphics[width=\linewidth]{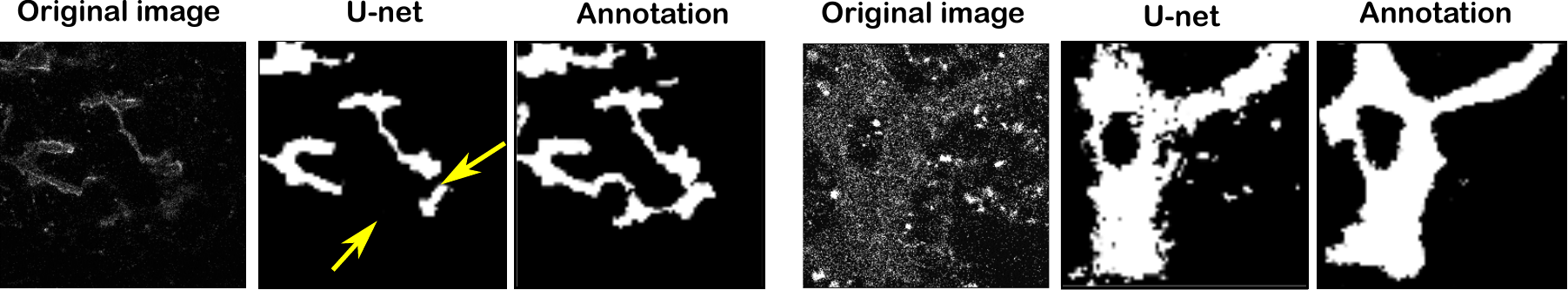}
\caption{Visualized results on 2D slices. \textbf{Left:} The arrows point to errors in the U-net prediction which are small when measured in voxel overlap, but large when viewed as differences in the corresponding tubular network structure. \textbf{Right:} Noise and artefacts lead to challenges in automatic segmentation of tubular structures.}
% left: label_ts_LI-2018-11-20-emb6-pos1_tp43-D1_D3_z8
% right: label_ts_LI-2018-11-20-emb6-pos1_tp104-D1_A1_z19
\label{panc}
\end{figure}

As can be seen from Figure~\ref{panc} (right), these images have a low signal-to-noise  ratio, which makes segmentation a difficult problem. Note that this is much lower than for other common tubular segmentation tasks, such as vessel segmentation. This is challenging not just because the images are noisy; also, the tubes appear at very different scales (compare the two images in Figure~\ref{panc}), and the image contrast is highest at the tubular boundary, meaning that tubes appear very different depending on their thickness relative to the image resolution as well as the depth in the image stack and local tissue depth.

Furthermore, most segmentation algorithms are trained to optimize voxel-based measures such as cross-entropy or accuracy. This, however, does not take into account how a small segmentation error viewed from the pixel point of view might, in fact, lead to a crucial error in the segmented network, see Figure~\ref{panc} (left).

\paragraph{The challenge of annotation.}
While unlabeled data is abundant, manual segmentation of 3D tubes is time-consuming and can only be performed by a trained expert. This growing divide between availability of data and availability of annotations is generic in biomedical imaging: While the cost of image acquisition and storage has decreased significantly, the cost of annotation remains the same, and learning useful representations from unlabeled data is thus desirable. 

As our application requires topologically accurate segmentations, it is also important to have examples of fully annotated 3D volumes in order to carefully assess the alignment of segmentations with annotations. As our training data consists of a few fully annotated 3D volumes, this means that the variation seen in the annotated training set is very limited.

\paragraph{Our contribution.} We propose a segmentation- and processing pipeline that allows the quantification of topological features (with an emphasis on network loops) of pancreatic tubular networks from noisy live imaging 3D microscopy. This pipeline consists of i) a semisupervised training strategy accompanied with a manual annotation strategy supporting it; ii) a novel and highly interpretable topological score function which quantifies the topological similarity between the ground-truth tubular network and the predicted one; iii) the use of multi-object tracking to postprocess detected loops with the help of temporal data, filtering out those loops that do not remain present over time.

In particular, our topological score can be used both to rank the topological consistency of different models, as well as to encourage topological consistency through hyperparameter tuning. We validate its use in both aspects against standard voxel-based model selection, as well as a topological loss function. We also compare with a segmentation network trained specifically with a topology-preserving loss function.

%+++++++++++++++++++++++++++++++++++++
\section{Related work}

\subsection{Topological consistency}
This is not the first work to consider topological consistency of segmentation algorithms. A series of previous works define segmentation loss functions that encourage topologically consistent segmentations. Some of these are based on computational topology, such as~\cite{hu:neurips:2019}, \cite{clough2020topological}, \cite{hu2020topology} or~\cite{wang2020topogan}. Such losses compare the numbers of components and loops between ground truth and segmentation throughout a parameterized ``filtration".  While appealing, this approach suffers from two problems: First, these algorithms come with a  high computational burden, which limit their application in typical biomedical segmentation problems: Early work on topological loss functions applied to 2D (or 2.5D) images, where computational limitations led to enforcing topological consistency of small patches rather than globally (\cite{hu:neurips:2019}). This is particularly problematic in our case, where the tubes appear on very different scales, making it hard to capture coarser scales with a single patch. This problem appears to persist with more recent work of \cite{hu2020topology} utilizing discrete Morse theory. The topological loss function introduced by \cite{clough2020topological} is computationally more feasible but requires the topology of the target object to be known a priori, even for test objects. This assumption holds for many tasks in bioimaging segmentation such as segmenting the 2D cross section of a tube or the boundary of a placenta imaged in 3D. However, this approach is not feasible for our setting, where we segment a full tubular structure with the goal of inferring its topology. For the same reason, methods such as~\cite{painchaud20} or~\cite{lee19}, which use conditional variational autoencoders or registered shape templates to constrain the topology of the segmented object, are also not applicable to our segmentation problem.

A second problem with methods based on computational topology- is that they tend to focus on topological equivalence of structure without taking its \emph{geometry} into account. More precisely, such methods tend to compare numbers of components and loops between prediction and ground truth without considering whether they actually \emph{match}. This is suboptimal if two patches contain non-matching components or loops; a problem which does occur in our data.

Another direction of research in topologically consistent loss functions, initiated by~\cite{shit2021cldice}, are based on soft skeletonizations which are compared between prediction and ground truth. We find these losses more applicable to our type of data, and compare to such a loss function in our experiments.

We empirically find that topological loss functions are not helpful for our segmentation problem, and we hypothesize that this is caused by the challenging nature of our data. We choose to take a step back and address these problems through model selection, for models trained with voxel based loss functions. As we are not using our score function to train the model, we can allow ourselves to incorporate non-differentiable components such as geometric matching of loops and components.  We thus introduce a topological score that measures topology preservation which is also geometrically consistent, in the sense that topology is represented by network skeletons, whose components and loops are soft matched based on geometry. This ensures that our score function is really assessing each topological feature on a global scale, and not just counting that their numbers match up. This also leads to a highly decomposable and interpretable score, which both decomposes into loop and component scores, and into precision and recall scores. The final topological score is defined based on their agreement.

\subsection{Semisupervised segmentation}
The problem of attaining annotations is even more severe in biomedical segmentation. As a result, there have been numerous attempts to make use of the information from unlabeled images to improve the segmentation task.

One line of thought is to enforce feature embeddings of neighboring labeled and unlabeled images to be similar to one another. The work of \cite{baur2019fusing} defines an auxiliary manifold embedding loss to cause labeled and unlabeled images which are close in input space to be also close in latent space. The authors of \cite{ouali2020semi} take a different approach to incorporating unlabeled data to learn better hidden representation features. 
%They first train an encoder and a decoder on labeled data, and then add in auxiliary decoders in addition to the original decoder for unlabeled data. The idea is to improve the shared encoder by ensuring auxiliary decoder outputs to be consistent with the original decoder output. 
They train a set of segmentation networks in the form of an encoder-decoder pair, where each network shares the same encoder. Only one of the decoders is trained on the labeled data, whereas all decoders predict a label map on the unlabeled data with the goal to minimize the distance between predictions. Variability between predictions is ensured via the introduction of random perturbations of the features generated by the encoder.
Another approach by \cite{zhang2017deep} uses adversarial networks. Along with a segmentation network, they also train a discriminator to distinguish between segmentation of labeled and unlabeled images. The authors in \cite{cui2019semi} add noise to the same unlabeled data as a regularization for a mean teacher model (\cite{tarvainen2017mean}).

A different semi-supervised approach is self-training (\cite{zoph2020rethinking}). In this approach, a segmentation model is trained on labeled images and then applied to unlabeled images to get pseudo-labels which are then added to the labeled pool for retraining. Generative adversarial networks (GAN) are used by \cite{hung2018adversarial} and \cite{mittal2019semi} to select better pseudo-labels by training a discriminator that distinguishes between pseudo-labels and ground-truth labels. Inspired by curriculum learning of \cite{bengio2009curriculum}, the work of \cite{feng2020semi} selects pseudo-labels gradually from easy to more complex. More recently, \cite{chen2021semi} have combined the consistency regularization with self-training. They train two networks with the same structure but different initializations, one for labeled data and one for unlabeled. Then they enforce consistency between the two networks while using the pseudo-labels from one network as supervisory signal for the other.

In our paper, we propose jointly training a segmentation network and an autoencoder in order to obtain features that generalize across a large, unlabeled training set. Other architectures have appeared in the literature that combine autoencoders and segmentation networks for regularization. This includes the anatomically constrained network of~\cite{oktay2018}, which utilizes an autoencoder to learn a global loss-function that enforces the network to produce anatomically realistic segmentations. Also related is the  segmentation network used by \cite{myronenko20183d}. In this work, a variational autoencoder (VAE) branch is added to their original encoder-decoder architecture with a purpose to regularize the common encoder network. While both of these archictectures bear similarities with ours, their use is different, as they apply the network to labeled data as a means to regularize the training. In our work, however, we utilize the autoencoder as a means to learn better latent space features that represent our entire dataset in a situation where the labeled part of the data is very scarce.

\section{Methods}

We start by discussing our semisupervised segmentation algorithm along with our baselines. Next, we design a novel and interpretable score function for assessing topological correctness of segmentation in a geometry-aware fashion. Finally, we introduce a topology-oriented post-processing scheme which utilizes temporal data and multi-object tracking to filter detected topological features depending on their presence over time.

\subsection{Semisupervised training of U-net}
\label{sec:architecture}
\begin{figure}[t]
    \centering
    \includegraphics[width=0.75\linewidth]{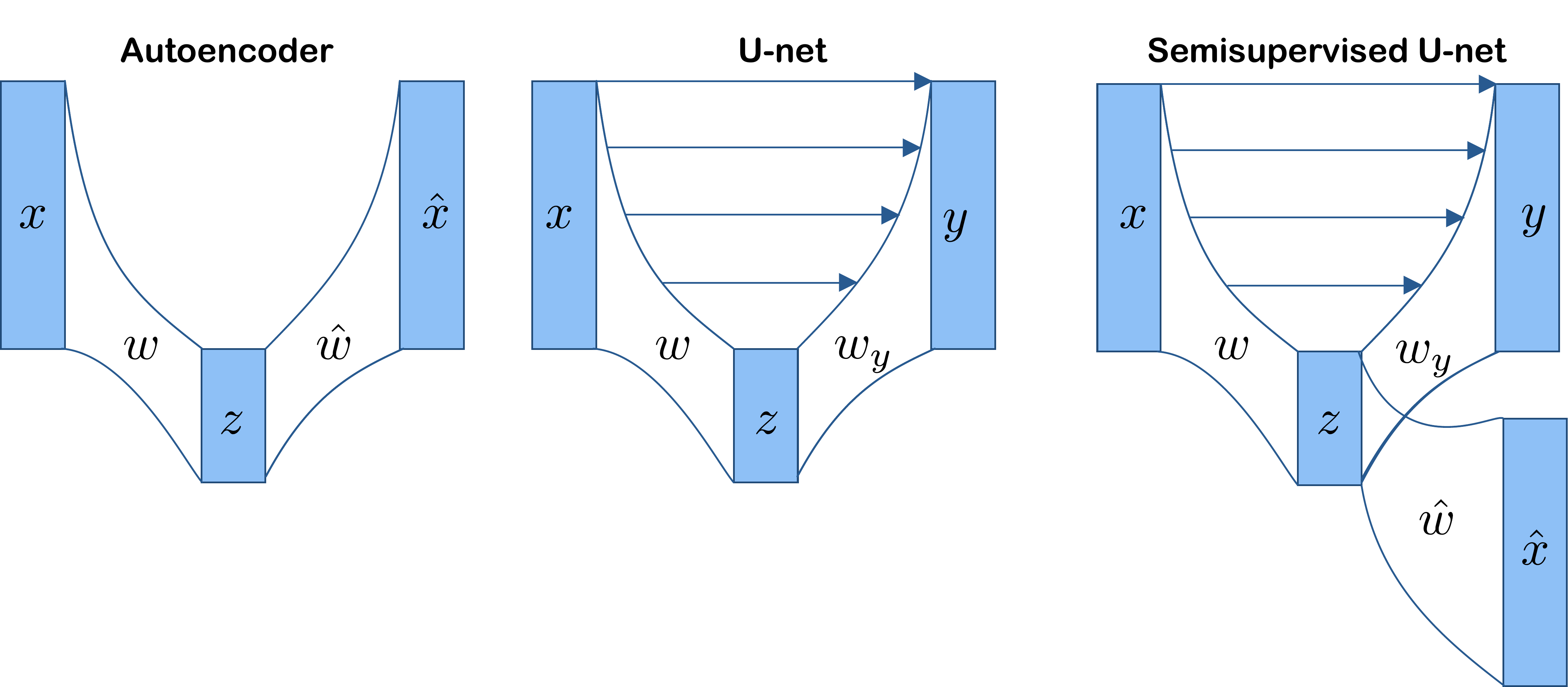}
    \caption{We apply semisupervised training of a segmentation U-net by combining an autoencoder (left) with a U-net (middle) into a combined network (right) which is simultaneously trained for segmentation and reconstruction. This allows us to learn features on a large, unlabeled dataset with rich variation, although we only have annotated segmentations for a small subset with little variation. In the architectures, $x$ denotes the input image, $y$ its corresponding binary segmentation map, and $\hat{x}$ is the reconstruction of the input image.}
    \label{fig:architectures_combined}
\end{figure}

Our application offers abundant unlabeled data and a small labeled subset of limited variation -- a very common situation in biomedical imaging. By learning features also from the unlabeled data, the increased diversity obtained by including unlabeled images might improve generalization to the full distribution of data expected at test time. To this end, we combine a 2D U-net (\cite{ronneberger2015u}) with an autoencoder, see Figure~\ref{fig:architectures_combined} (right). We refer to this as a semisupervised U-net.

The loss function for the semisupervised U-net is the weighted combination $L = L_R + \alpha L_S$ of the reconstruction loss $L_R$ and the segmentation loss $L_S$ where $\alpha$ is a hyperparameter. In our experiments, $L_R$ is the mean squared error, and $L_S$ is binary cross-entropy. The segmentation loss is set to zero for unlabeled images. Although the segmentation decoder $w_y$ is trained only on the labeled images, it is implicitly affected by the unlabeled data via its dependence on the encoder $w$ which, along with $\hat{w}$, is trained on both labeled and unlabeled images.
% \begin{SCfigure}
%     \includegraphics[width=0.5\linewidth]{weights.pdf}
%     \caption{The encoder, reconstruction and segmentation networks are represented by their weights $w$, $\hat{w}$ and $w_y$, respectively. Every conv and up-conv is followed by a ReLU function.}
%     \label{fig:weights}
% \end{SCfigure}

We compare the semisupervised U-net framework to i) a U-net initialized randomly and trained solely on labeled images and ii) a U-net trained on labeled images, whose encoding weights $w$ are initialized at the optima of an autoencoder trained on both labeled and unlabeled images. What separates the semisupervised U-net architecture from the U-net pre-trained on an autoencoder is the joint optimization of the two losses. We believe---as backed up by our experiments---that this joint optimization tailors the representations we learn from the unlabeled images to suit the segmentation task, and help them generalize to more diverse images than those annotated for training.

\subsection{Topological score function}
\label{sec:toposcore}

In order to perform model selection that prioritizes topologically consistent segmentations, as well as to evaluate our ability to correctly detect topological features such as loops and components, we design a topological score function. While existing topological loss functions often focus on getting the correct number of topological features such as components and loops, it is extremely important for our application to detect not only the correct number of loops, but the correct loops. In order to quantify this, our topological score therefore relies on first matching loops and components, and then measuring their correctness via overlap. In this way, the score function is also geometry-aware.

\begin{figure}[t]
    \centering
    \includegraphics[width=0.8\linewidth]{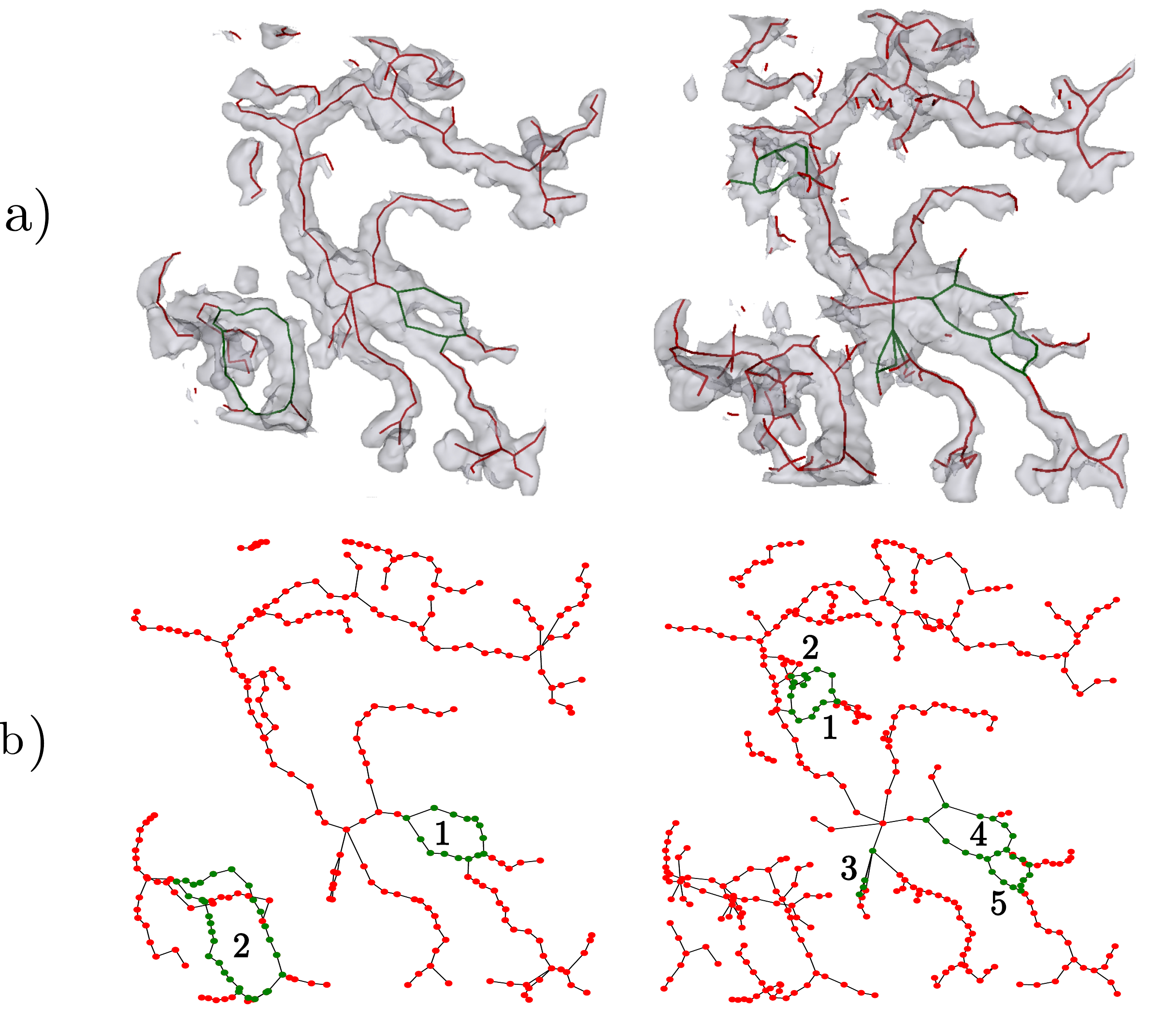}
    \caption{Ground truth (left) and prediction (right): a) segmentation and their skeletons. b) 2D projections of the skeletons onto the $(x,y)$-plane shown as nodes and edges. Green lines in a) and green nodes in b) take part in loops. Note that the ground truth segmentation has 2 loops and 7 components, while the prediction has 5 loops and 10 components. The loops are identified by numbers, for the sake of the discussion below.}
    \label{fig:skeleta}
\end{figure}

The loop- and component matching problem has some important differences from the graph matching problems widely covered in the literature. Our loops and components are retrieved via skeletonization of the manual segmentation and the estimated segmentation, respectively. We expect these skeleta to be qualitatively different: As the manual segmentations are more smooth than the estimated ones, their skeleta have significantly fewer nodes. This is illustrated in~Figs.~\ref{fig:skeleta} and \ref{fig:example_res} One consequence of this is that traditional graph matching algorithms, which are typically designed to make a series of local compromises in order to optimize a global objective, would be very difficult to tune. Instead, we take advantage of knowing that  the ground truth- and estimated skeleta both represent the same underlying geometric tubular system. We therefore expect these skeleta to be spatially close together, and we use local correspondences between points on the two skeleta to obtain loop- and component matching, as detailed below.

The topology of a segmented tubular structure is represented via its skeleton. We utilize a robust skeletonization algorithm recently proposed by \cite{baerentzen2020skeletonization} and then use NetworkX of \cite{hagberg2008exploring} to identify the loops and components, see Figure \ref{fig:skeleta} for an example. We design our topological score via the following steps: i) matching skeletal nodes; ii) soft-matching components and loops using a point cloud Intersection-over-Union (IoU) score; and iii) collecting these into a topological score. By matching topological features based on geometric affinity, we ensure that our measured topological consistency is also \emph{geometrically} consistent, meaning that the IoU score considers overlap of geometric features that appear near each other.

\begin{figure}[t]
    \centering
    \includegraphics[width=\linewidth]{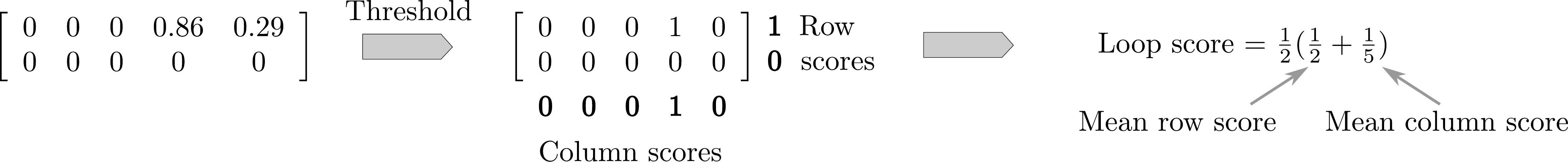}
    \caption{Example computation of the loop score for the prediction shown in Figure~\ref{fig:skeleta}. The ground truth segmentation has 2 loops while the prediction has 5. Here, loop \#1 in $S_{gt}$ has overlap with loops \#4 and \#5 in $S_p$, where the first is thresholded as a perfect match and the second is thresholded as no match. The row score = [1, 0] means loop \#1 from the ground truth is very confidently matched to a predicted loop; the column score = [0, 0, 0, 1, 0] means that the loop \#4 from prediction is very confidently matched to a loop in ground-truth.}
    \label{fig:toposcore}
\end{figure}

For the example shown in Figure \ref{fig:skeleta}, each step explained below is illustrated in Figure \ref{fig:toposcore}.

\begin{figure}[t]
    \centering
    \includegraphics[width=0.9\linewidth]{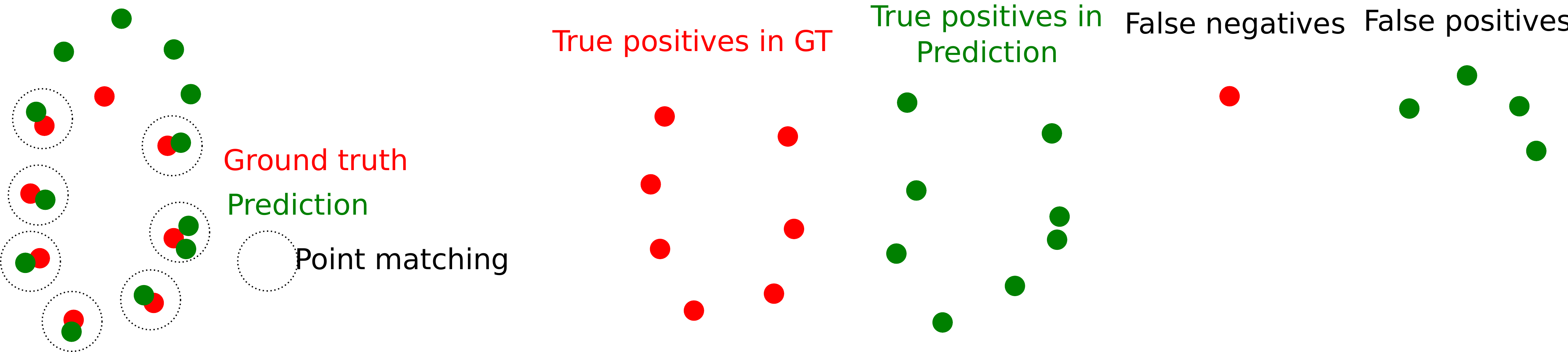}
    \caption{An example of point matching for two loops (edges not drawn).}
    \label{fig:IoU_points}
\end{figure}

\paragraph{Step i) Matching individual nodes.} We denote the skeleton graphs of ground truth and prediction by $S_{gt}$ and $S_p$, respectively. Due to noise, predicted segmentations are bumpier than the ground truth, giving higher node density in $S_p$ than in $S_{gt}$. Examples of this phenomenon can be seen in Figs.~\ref{fig:skeleta} and~\ref{fig:example_res}. The topological score needs to be robust to this. As a first step towards creating a density-agnostic score, we therefore match skeletal components and loops by first matching their nodes as follows (see also Figure~\ref{fig:IoU_points}): 

For every loop in one of the skeletons $S_{gt}$/$S_p$, each node is matched to its nearest node in a loop from the \emph{opposite} skeleton $S_p$/$S_{gt}$ if such a node exists within a fixed radius $r > 0$; otherwise the node is considered unmatched. Matching several nodes in one skeleton to the same node in the other skeleton is allowed, as can be seen from Figure~\ref{fig:IoU_points}. 

The same matching procedure is applied to components. 

%Note that the matching is asymmetric -- even if the node $n_{gt}$ in $S_{gt}$ is matched to the node $n_p$ in $S_p$, it is possible that $n_p$ is not matched with $n_{gt}$. 

\paragraph{Step ii) Soft matching of loops and components via Point Cloud IoU.} Next, we perform partial matching of components and loops on nodes of the 3D skeleta of ground-truth and prediction. With skeleton nodes matched, some components or loops may overlap partly, as shown in Fig~\ref{fig:IoU_points}. We need to measure this, again being robust to differences in skeleton node density.

Define $\TP_{p}$ to be the number of nodes in the predicted loop that is matched to a point in the ground truth (GT) loop; $\TP_{gt}$ the number of nodes in the GT loop that is matched to a point in the predicted loop, $\FP$ the number of nodes in the predicted loop that is not matched to a point in the GT loop, and $\FN$ the number of loops in the GT loop that is not matched to a point in the predicted loop. Similarly for components. Now, we might naïvely define the loop's IoU score as

\[
\frac{\TP_{gt} + \TP_{p}}{\TP_{gt} + \TP_{p} + 2\FP + 2\FN}.
\]

However, this is not robust to differences in node density, as a very dense, but incomplete, prediction would get an inflated score. Instead, in analogy with samples from probability distributions, we define a point cloud IoU, which we use to measure loop-wise performance:

\begin{equation}
\IoU = \cfrac{\cfrac{\TP_{gt}}{\TP_{gt} + \FN} + \cfrac{\TP_{p}}{\TP_{p} + \FP}}{\cfrac{\TP_{gt}+ 2\FN}{\TP_{gt} + \FN} + \cfrac{\TP_{p} + 2\FP}{\TP_{p} + \FP} }~,
\label{iou}
\end{equation}
 where $\TP_{gt}$ and $\FN$ have been normalized by the number of nodes in the ground truth loop ($\TP_{gt}+\FN$), and $\TP_{p}$ and $\FP$ have been normalized by the number of nodes in the predicted loop ($\TP_{p}+\FP$). 
 
Our score function depends highly on the matching radius $r$. As $r$ is increased, it is more likely for nodes between distant loops or components to become matched and consequently, FP and FN become smaller. Thus, with large matching radius, $\IoU$ will become one. This makes intuitive sense, as large $r$ indicate that different skeletonizations of the same structure have large spatial variability and this must be taken into account by the matching. A consequence of this is that $r$ can only be chosen by hand, based on visual inspection and hand-annotation of matching pairs between ground-truth and prediction.

\paragraph{Step iii) A topological score for comparison of segmentations.} Given skeletons $S_{gt}$ and $S_p$, we separate our measure of topological consistency into performance for loops and performance for components. To this end, we define two topological scoring matrices; one for components and one for loops. As in the example of Figure~\ref{fig:toposcore}, the loop score matrix has size $L_{gt} \times L_p$, where $L_{gt}$ and $L_p$ is the number of loops in $S_{gt}$ and $S_p$, respectively. For each pair of GT and prediction loops, their matrix entry is given by their point cloud IoU of Eq.~\eqref{iou}; this completes the matrix on the left side of Figure \ref{fig:toposcore}. Any IoU below a threshold $t_{low}$ is set to 0, indicating no match, and any IoU over $t_{high}$ is set to 1, indicating perfect match. If an IoU is between $t_{low}$ and $t_{high}$, then it is left unchanged in the matrix. (Figure \ref{fig:toposcore} middle matrix)

The thresholds $t_{low}$ and $t_{high}$ are necessary because without such a thresholding, we easily lose the distinction between many poor matches (with small but nonzero contribution to the score) and a single good match (with decent but not maximal contribution to the score). Without such a thresholding, even perfect matches will not give scores near 1 because of the dependence on the imperfect skeleta.

By adding horizontally we obtain a row score indicating the degree to which a ground truth loop is matched to a loop in the predicted topology. Similarly, summing vertically, the column score indicates the degree to which a predicted loop is matched to a loop in the ground truth topology. The elements in these row and column scores can, in principle, add up to more than one when nodes appear in multiple loops; they are clipped to be at most $1$.

We can now take the mean of row scores and column scores to draw an analogy to confusion matrix terminology. The mean row score is similar to recall, rewarding fewer false negatives; the mean column score is similar to precision, rewarding fewer false positives; and the mean of the two mean scores is similar to the F1 score; we refer to this as the loop score (Figure \ref{fig:toposcore} right side). Note that in this way, the topological score is decomposable into a series of highly interpretable sub-scores, which makes it both interpretable and adaptable to a wide range of applications.

A component score is defined in the same manner as the loop score. The final topology score is given by the mean of the loop and component scores.

The above does not handle cases where either $S_{gt}$ or $S_p$ have no loops (or components). If neither skeleton has loops, the loop score is 1. If $S_{gt}$ has no loops, then the loop score is ${1}/({1 + \# err})$, where $\# err$ is the number of loops found in $S_p$. If the $S_p$ has no loops, then the final score is ${1}/({1 + \# err})$, where $\# err$ is the number of loops found in $S_{gt}$. Components are handled similarly.

\subsection{Postprocessing of loops using temporal information}

In this section, we use the temporal structure of our data, which consists of consecutive 3D images taken over time, to filter out spurious loops as a postprocessing step. To do so, we first need to track loops, and then remove those trajectories which last a very short time.

We track loops by tracking their centers, and treat this as a multi-object tracking problem, whose validation can be broken into two questions: 1. Are predicted loop centers matched to a loop center in the ground-truth? 2. Do loop centers in different frames belong to the same loop? The first question is known as detection and the second as association. It is important to emphasize that the association of loops is implicitly affected by the detection of loops, which has been one of the topological features we have been interested in.
    
HOTA (\cite{luiten2021hota}) is a multi-object tracking metric capturing both the detection accuracy (DetA) and the association accuracy (AssA). The final HOTA metric is defined as the geometric mean of DetA and AssA. The detection accuracy is based on a bijective matching between predicted- and ground-truth loop centers, where the distance between a matched pair cannot exceed a fixed threshold $\gamma$. To quantify association accuracy given predicted- and ground-truth trajectories of loop centers, \cite{luiten2021hota} propose a measure which captures how accurate the alignment of predicted- and ground-truth trajectories is for every matched center. Finally, both detection- and association accuracies can be further broken down to recall (Re) and precision (Pr) constituents, i.e. DetRe, DetPr, AssRe, AssPr, which are useful for interpreting the results.
    
We use Trackpy (\cite{allan_daniel_b_2021_4682814}) to first track predicted loop centers,and next remove trajectories which last a short time. In order to do that, Trackpy includes a few hyperparameters: \textit{search range} (the maximum distance loop centers can move between frames), \textit{adaptive stop} (the value used to progressively reduce search range until solvable), \textit{adaptive step} (the multiplier used to reduce search range by), \textit{memory} (the maximum number of frames during which a loop can vanish, then reappear nearby, and be considered the same loop), and \textit{threshold} (the minimum number of frames a feature has to be present in order not to be removed). These hyperparameters are tuned using the HOTA metric by grid-search on movies for which we have target trajectories annotated.
    
For components, we take a different approach as tracking component centers is not reasonable. At every time point, we compare all components on the current frame to all components from consecutive frames. Given a fixed component from the current frame, if there is no component in its `vicinity' from neighboring frames then that particular component is removed. We use the minimum Euclidean distance between the nodes of two components as a measure of distance between them, and if this distance is lower than a threshold then the two components are in the same vicinity.
	
\section{Data and preprocessing}
	
The data used in this paper is extracted from 3D images obtained by live imaging of embryonic mouse pancreas (E14.0) explants. An image is recorded every 10 minutes during a period of 48 hours using a Zeiss LSM780 confocal microscope equipped for live imaging. The image resolution is $0.346 \times 0.346 \times 1.0  ~\mu m \textrm{ (x, y, z)}$, and the dimension size is $1024 \times 1024 \times D$ pixels, where $D$ varies between 27 and 40.

\begin{figure}[t]
    \centering
    \includegraphics[width=\linewidth]{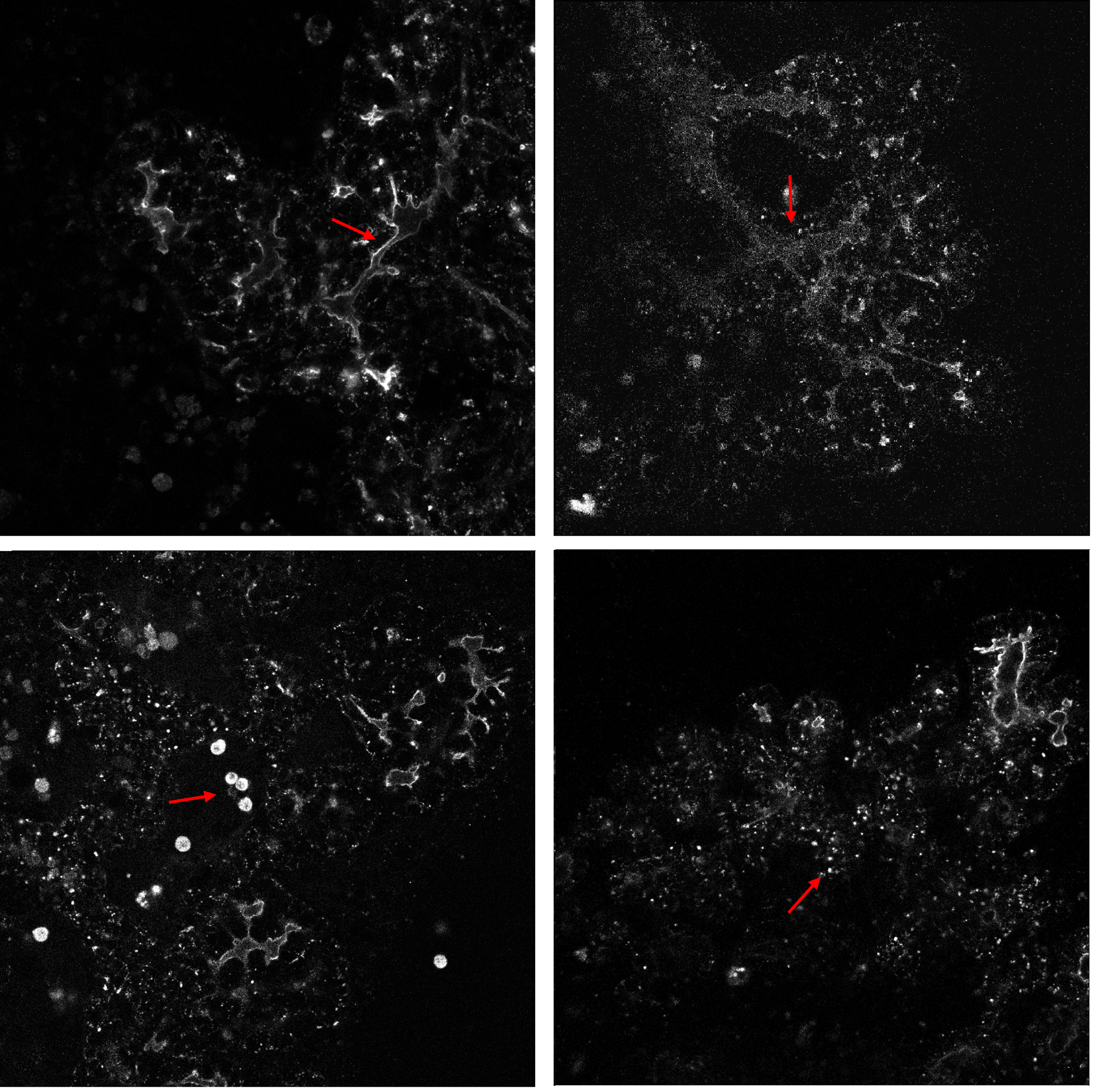}
    \caption{2D slices from different movies extracted to show different image artifacts. \textbf{Top left:} a case where the boundary of tube has a high intensity. \textbf{Top right:} a case where the intensity of boundaries are no higher than the rest of the tube. \textbf{Bottom left:} large bright spots due to red blood cells and other dead cells. \textbf{Bottom right:} small bright spots due to the reporter protein in cells being transported to the tubular surface.}
    \label{fig:artifact}
\end{figure}

The imaged mouse pancreas expresses a reporter gene leading to production of a red fluorescent protein (mCherry), which is localized to the cell's apical side facing the tube's inner surface. In the images, we see the fluorescence as a bright tubular boundary surrounding a dark inner lumen, see Figure~\ref{fig:artifact}. Due to biological variation between individual samples, the fluorescence intensity is lower in some data sets, leading to a lower intensity boundary. In addition, low fluorescence is seen at the furthest end of the z-dimension due to intensity decay through the z-dimension. An additional segmentation problem is presented by the fact that the tubes have very varying diameters: Some tubes are so wide that the inner dark space can be mistaken for space between tubes. In other cases, segmentation is hard because the tube is so narrow that the dark inner space is hard to detect. These effects lead to a challenging tubular segmentation problem.

The problem is further complicated by several common artifacts (Figure~\ref{fig:artifact} bottom row): First, we see imaging artifacts in the form of big clusters of very bright pixels in some images. These are caused by red blood cells and dead cells which fluoresce in the same wavelength as the reporter. There are also small clusters of bright pixels, caused by the production of reporter protein inside the cells, that are being transported to the tubular surface. These are mainly located at the end of the z-dimension where the cells attach to the dish and move significantly over time, which is helpful in ignoring them.

\paragraph{Intensity normalization} The mean intensity is close to zero, as the foreground volume is relatively small. The intensity variance, however, differs considerably across images, in particular due to the bright spot artifacts mentioned above. As preprocessing, every image is standardized to have zero mean and standard deviation $1$, and image intensities are clipped to take values within $3$ standard deviations.

\paragraph{Annotation strategy} As the image content can vary a great deal within a single image, each 3D image which was later fully or partially annotated, was first divided into a $4 \times 4$ grid of patches in the $x$-$y$ plane, each of size $256 \times 256 \times D$. Those patches that had no discernible tubular structures were labeled as such and left out from the manual annotation and downstream analysis. We refer to these patches as ``low information patches'' below.

For the training set, $6$ full 3D images from $3$ different movies were manually segmented ($2$ 3D images from each movie). The choice of annotating a few full 3D volumes for training was made in order to have some manually segmented examples available early on in which a complete tubular network was visible. For the
sake of validating and testing on as diverse a dataset as possible, 10 3D patches were selected for validation, and 20 3D patches were selected for testing, in which the tubular structure was annotated by the laboratory assistant. There is no overlap in tissue samples between the test split and the training/validation splits.

\paragraph{Training on patches} All our models are trained on image patches. Labeled training patches were obtained by drawing them at random from the $6$ fully annotated 3D images; these patches were restricted not to overlap with those patches labeled as low information patches. The training set for the autoencoder and semisupervised U-net further includes patches from 20 unannotated images from 10 movies, also referred to as unlabeled training set. 

To avoid artifacts and degradation of performance near patch boundaries, we pad the image patches with a 32 pixel wide border, making boundary effects negligible. We thus use $320 \times 320$ patches, extracting predictions for the inner $256 \times 256$ region.

\paragraph{Annotated loops} Our ultimate goal is to robustly detect loops in the tubular structure. To assess our ability to do so, loops were manually tracked and annotated over time for a small number of movies. More precisely, the $x$- and $y$- coordinates of loop centers were recorded in 5/1/3 movies from the unlabeled training set/validation set/test set, respectively. Moreover, the loops detected by our proposed pipeline (postprocessed output of semisupervised U-net) were assessed by a trained expert on 4 test movies.

\section{Experiments and results}
In Table~\ref{tab:res}, we compare three objectives for hyperparameter selection, i.e. voxel IoU, clDice measure, and the proposed topology score in this paper, and apply them to three models, namely a fully supervised 2D U-net, a U-net pre-trained on an autoencoder (AE+U-net), and the semisupervised U-net. Moreover, we take the same fully supervised U-net architecture and compare it to the case where the topology-preserving loss of \cite{shit2021cldice} has been used during training (clDice U-net).

\begin{figure}[t]
\centering
    \includegraphics[width=0.9\linewidth]{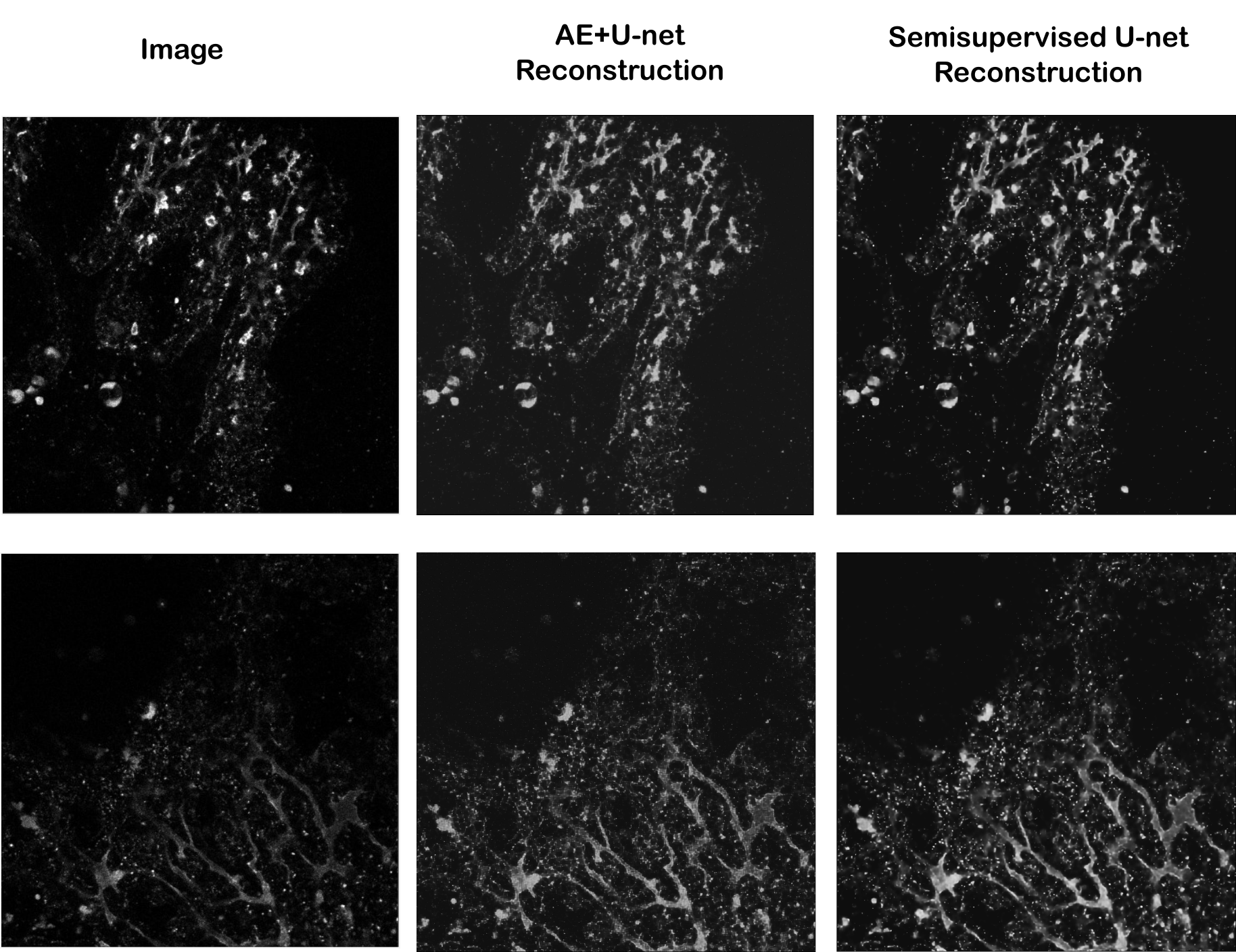}
    \caption{Two 2-dimensional slices from the test set (left column) and their corresponding reconstruction at the final epochs of AE+U-Net (middle column), and semi-supervised U-net (right column).}
    \label{fig:rec}
\end{figure}

\paragraph{Training setting} All models are optimized using Adam with $\beta_1 = 0.9$ and $\beta_2 = 0.999$. The learning rate for all architectures is $10^{-5}$, except for clDice U-net which is set to $10^{-2}$ for better training; these were selected from the values $\{10^{-2}, 10^{-3}, 10^{-5}\}$ based on the robustness of the training- and validation set learning curves. Random weight initializations are done by Xavier uniform initializer, and biases are set to zero initially. The U-net and clDice U-net are trained for 200 epochs. For the AE+U-net, the autoencoder  was trained for 20 epochs, which resulted in satisfactory reconstructions (Figure \ref{fig:rec}), and its U-net part was trained for 200 epochs. For the semisupervised U-net, the number of epochs was set to a 40, as it contains both labeled and unlabeled data. The hyperparameter $\alpha$ in the combined loss of the semisupervised U-net was selected from the range $\{0.1, 1, 10, 50\}$ by tuning performance on the validation set, obtaining $\alpha=10$. The weighting between soft-Dice and clDice in the loss function of clDice U-net is set to be equal.

\begin{table}[t]
\caption{Segmentation performance assessed via voxel IoU, clDice measure, the topological score, and its sub-components, the loop score and the component score. (Sec.~\ref{sec:toposcore}). For each of the 3 models, the best-performing threshold (Thr) was selected according to Voxel IoU, clDice measure, and Topological Score on the validation set. Numbers in bold indicate the highest mean across both architectures and tuning criteria. Note that the three rows for "Semisupervised U-net" are identical because all tuning criteria chose the same segmentation threshold.\\
The last row shows the performance of ``clDice U-net", a U-net model trained with the topology-preserving loss ``clDice" of~\cite{shit2021cldice}. As the pixel-wise probability values of this model clustered as either less than 0.1 or more than 0.9 on the validation set, we chose 0.5 as the threshold to apply on the test set.}
\resizebox{\textwidth}{!}{%
      \begin{tabular}{ccc|ccc|cc}
    \hline
      Architecture & Tuning Criterion & Thr & Voxel IoU & clDice & Topological Score & Loop Score & Component Score\\
      \hline
       \multirow{3}{*}{U-net} & Voxel IoU & 0.8 & 0.474±0.208 & 0.535±0.192 & 0.395±0.263 & 0.312±0.297 & 0.477±0.285\\ 
       & clDice  & 0.8 & 0.474±0.208 & 0.535±0.192 & 0.395±0.263 & 0.312±0.297 & 0.477±0.285\\
       & Topology Score  & 0.9 & 0.449±0.214 & 0.529±0.212 & 0.424±0.265 & \textbf{0.390±0.339} & 0.458±0.297\\
       \hline
            \multirow{3}{*}{AE+U-net} & Voxel IoU & 0.5 & 0.447±0.210 & 0.504±0.196 & 0.356±0.243 & 0.308±0.299 & 0.404±0.264\\
             & clDice  & 0.6 & 0.447±0.207 & 0.504±0.206 & 0.382±0.264 & 0.333±0.320 & 0.430±0.279\\
      & Topology Score  & 0.7 & 0.418±0.214 & 0.488±0.219 & 0.375±0.222 & 0.325±0.311 & 0.425±0.271\\
      \hline
      \multirow{3}{*}{Semisupervised U-net} & Voxel IoU & 0.7 & \textbf{0.490±0.226} & \textbf{0.555±0.213} & \textbf{0.436±0.253} & 0.386±0.352 & \textbf{0.486±0.271}\\ 
       & clDice  & 0.7 & \textbf{0.490±0.226} & \textbf{0.555±0.213} & \textbf{0.436±0.253} & 0.386±0.352 & \textbf{0.486±0.271}\\
      & Topology Score & 0.7 & \textbf{0.490±0.226} & \textbf{0.555±0.213} & \textbf{0.436±0.253} & 0.386±0.352 & \textbf{0.486±0.271}\\
     \hline
      & clDice U-net & 0.5 & 0.449±0.193 & 0.443±0.187 & 0.337±0.263 & 0.314±0.346 & 0.360±0.275\\
    \hline
  \end{tabular}}
  \label{tab:res}

\end{table}
´

\begin{figure}[t]
    \centering
    \includegraphics[width=\linewidth]{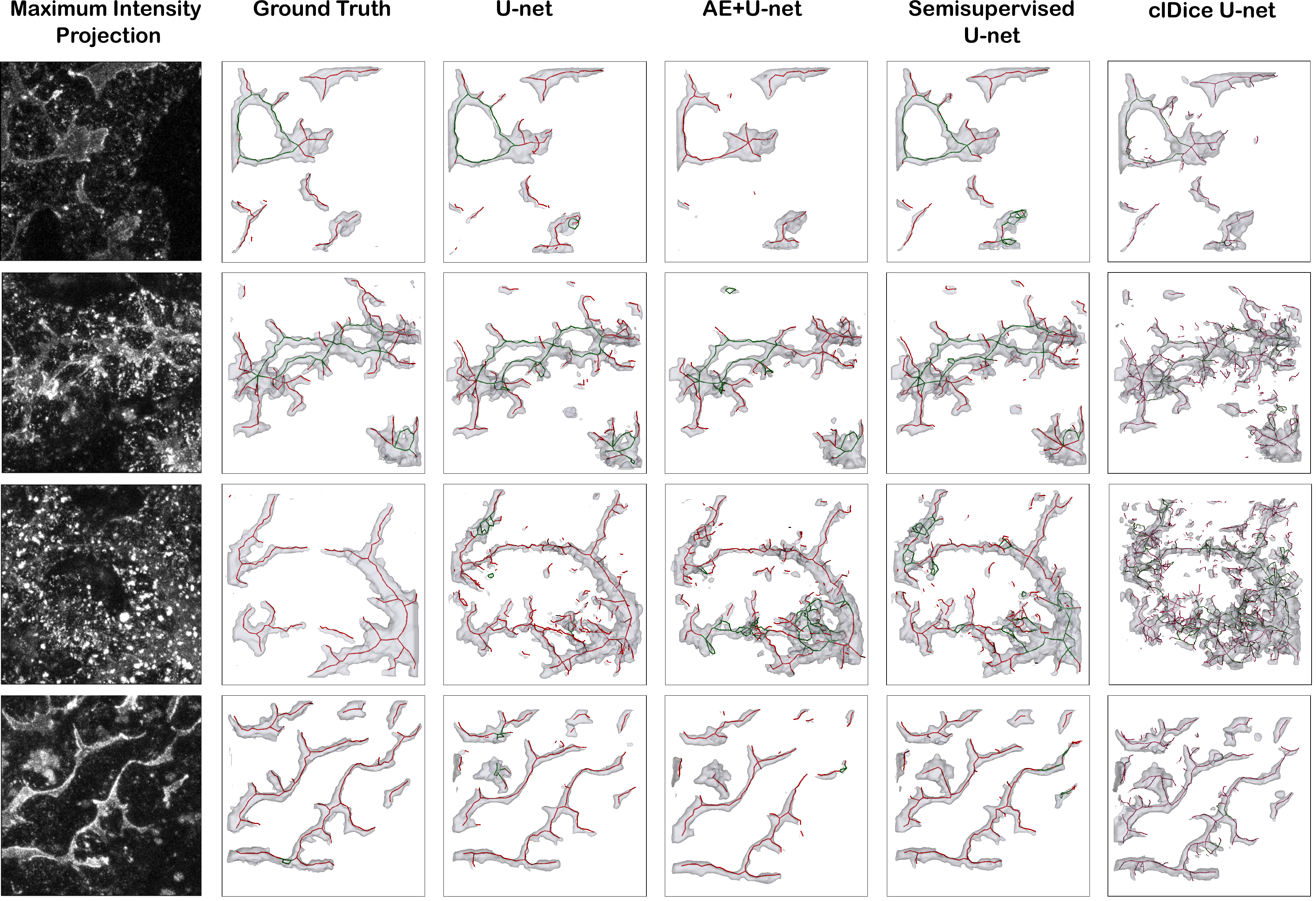}
    \caption{Segmentation results for 4 example test patches.}
    \label{fig:example_res}
\end{figure}

\subsection{Segmentation performance} \label{sec:tuning}
The hyperparameter that we tuned in our experiments was the cutoff threshold applied to the outputs of the sigmoid function to make a binary segmentation. The thresholds of interest ranged from 0.1 to 0.9 with a 0.1 increment. The thresholds performing best on the validation set were applied to the test set, and then skeletons were computed using the robust GEL skeletonization algorithm (\cite{gel}). Skeleton components of size $< 5$ nodes were ignored; the radius $r=10$ pixels was used for node matching; and the thresholds $t_{low} = 0.3$ and $t_{high} = 0.7$ were used for the component and loop score matrices. The radius $r$ was chosen by visual inspection (using the training- and validation sets) of differences between ground truth- and estimated skeleta on the validation set. The thresholds $t_{low}$ and $t_{high}$ were similarly set by visual inspection on the validation set of those loop matching being thresholded to the values $1$ or $0$.

\begin{figure}
    \centering
    \includegraphics[width=0.9\linewidth]{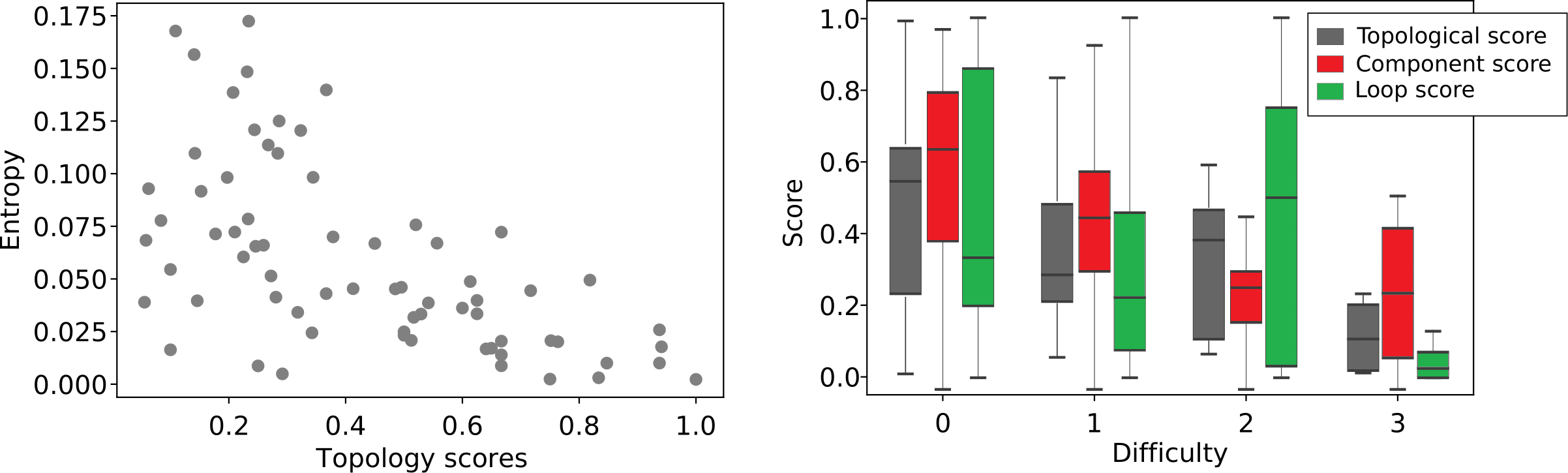}
    \caption{\textbf{Left}: Patch-wise topology score versus segmentation entropy. \textbf{Right}: Distribution of topology, component and loop scores divided over difficulty of manual labeling.}
    \label{fig:difficulty}
\end{figure}

In Table~\ref{tab:res}, we report mean and standard deviation of patch-wise scores on the test set. Note that this standard deviation indicates variation in performance over different patches, not robustness over multiple training runs. It is worth noting that, given a fixed model, all performance measures are only dependant on the threshold value. As a result if multiple tuning criteria reach the same threshold value for a model, all performance measures for those tuning criteria would be identical. For visual comparison, Figure~\ref{fig:example_res} shows four randomly selected patches from the test set to compare the ground truth annotation with different model predictions at their best-performing thresholds according to the topological score. As we believe the difficulty of the segmentation problem varies greatly among different patches, in Figure~\ref{fig:difficulty} we have plotted topology score against entropy (left) and distribution of topology-, component- and loop scores for each level of difficulty, evaluated by an expert. Both plots use the test set predictions of the semisupervised U-net.

\subsection{Loop tracking and filtering}
\label{sec:loop}
We use the temporal data to remove false positives from the detected loops. Due to computational constraints, we only apply this to the best performing model, i.e the semisupervised U-net, and the clDice U-net.

The hyperparameters of Trackpy, used to track loops as well as filter them, were tuned on 6 movies with annotated loops. Of these, 5 came from the unlabeled training set and 1 came from the validation set. The resulting optimal hyperparameters are found in Table \ref{tab:trackpy}. %(search range = 15), (adaptive stop = 5), (adaptive step = 0.9), (memory = 1), and (threshold = 15).
We have used Euclidean distance to match loop centers, and the upperbound for the distance in HOTA measure is $\gamma = 50$ pixels.

After filtering, the loop scores on the test set patches were improved from \textbf{0.386±0.352} to \textbf{0.808±0.281} for the the semisupervised U-net and from \textbf{0.314±0.346} to \textbf{0.762±0.301} for the clDice U-net.

\begin{figure}[t]
\centering
\includegraphics[width=0.9\linewidth]{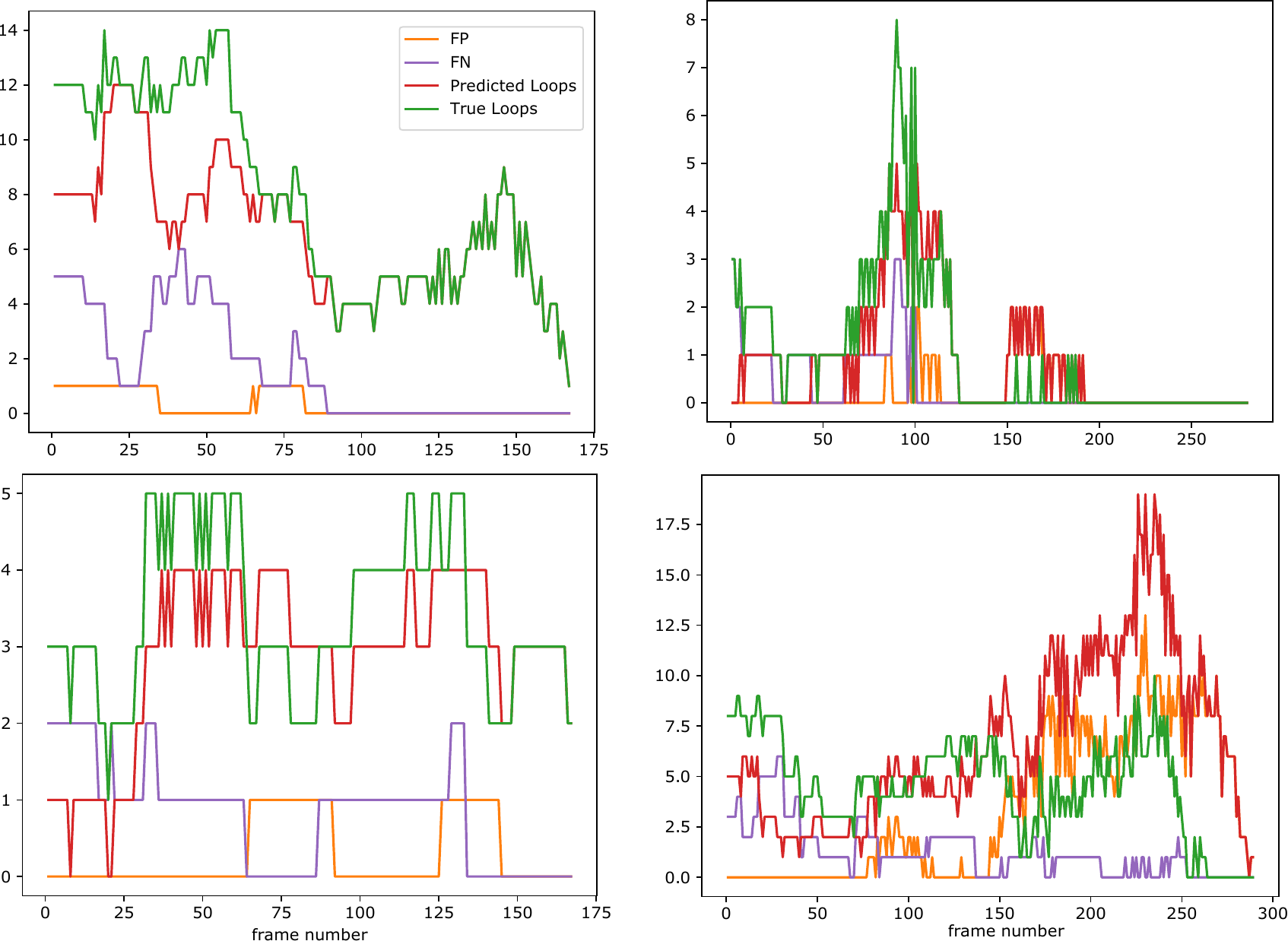}
\caption{Each plot is a movie from the test set showing the performance of loop detection evaluated by an expert. The y-axis measures the number of false positives (FP), false negatives (FN), predicted loops, and true loops.}
\label{fig:pia}
\end{figure}

These final results were also evaluated by a trained expert, who annotated true positive (TP), false positive (FP), and missing (false negative, or FN) loops. This was carried out for 4 full movies from the test set; see Figure \ref{fig:pia}. As can be seen from the results, there is a strong correlation between predicted and true loops. In the bottom right movie, we see a large number of false positives; this movie was also annotated as challenging by the trained expert due to the high number of loops and increase in noise towards the end of the movie.

Table \ref{tab:hota} shows the HOTA scores and sub-scores for the 3 movies in the test set for which we have had manual tracking of loop centers.

\begin{table}[t]

\caption{Parameters of tracking and filtering loop centers in Trackpy for two architectures.}
\label{tab:trackpy}
\centering
      \begin{tabular}{|c|c|c|c|c|c|c}
    \hline
      Architecture & search range & adaptive stop & adaptive step & memory & threshold\\
      \hline
      Semisupervised U-net & 15 & 5 & 0.9 & 1
      & 15\\
    clDice U-net & 10 & 5 & 0.9 & 1 & 12\\
       \hline
  \end{tabular}
\end{table}

\begin{table}[t]
\caption{Performance of tracking loop centers for two architectures.}
\label{tab:hota}
\centering
      \begin{tabular}{|c|c|ccc|ccc|}
    \hline
      Architecture & HOTA & DetA & DetRe & DetPr & AssA & AssRe & AssPr\\
      \hline
      Semisupervised U-net & 0.337 & 0.256 & 0.361 & 0.649
      & 0.469 & 0.478 & 0.956\\
    clDice U-net & 0.275 & 0.233 & 0.279 & 0.705
      & 0.324 & 0.330 & 0.944\\
      \hline
  \end{tabular}
  \end{table}

\begin{figure}
    \centering
    \includegraphics[width=\linewidth]{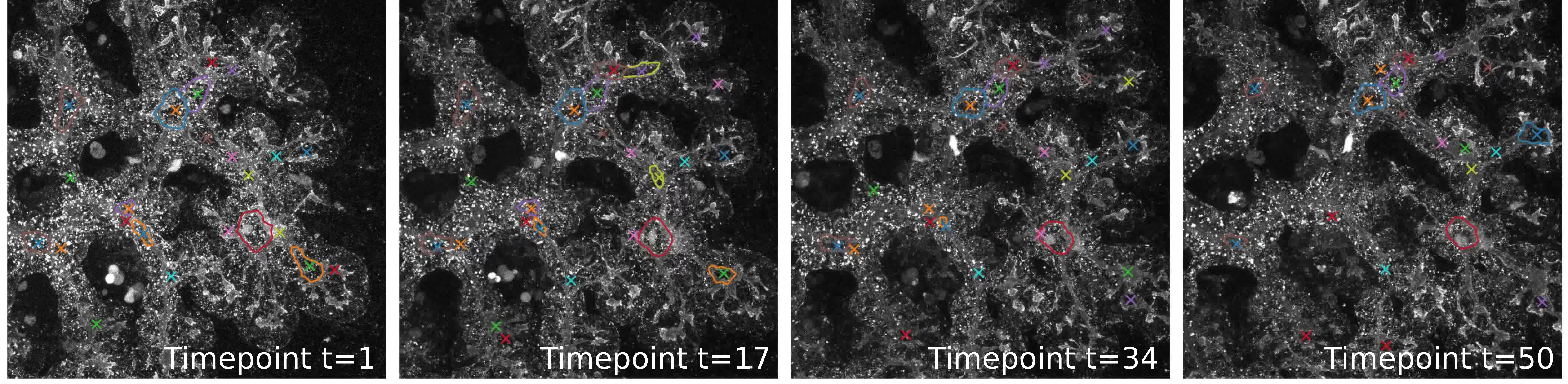}
    \caption{Frames from one of the test set movies with loops manually annotated by an expert shown as crosses. The predicted loops are shown as colored loops. Note that most of the mistakes made are false negative, i.e.~not detecting a loop that was annotated by the expert. This supports the use of the detected loops for testing biological hypotheses, in the sense that the detected loops are likely to be real.}
    \label{fig:movie}
\end{figure}
  
\section{Discussion and Conclusion}

We have tackled the challenging problem of segmenting pancreatic tubes from live-imaging microscopy, in order to track loops in the complex and dynamic tubular network that they form. This came with three important challenges: 1) Having a topologically accurate segmentation in order to correctly track the loops of the tubular network; 2) Utilizing temporal data to postprocess the segmentation results; and 3) Utilizing unlabeled data to aid the segmentation in generalizing to diverse test data, as our annotated training data has very limited variation.

To obtain topologically accurate segmentations, we derive a topological score, which splits into loop- and component scores, to quantify the topological correctness of the segmentations. To aid generalization, we train a segmentation U-net in a semisupervised fashion, combining an autoencoder reconstruction loss applied to a large set of un-annotated images, with a segmentation loss applied to the annotated images. Finally, we make use of large amounts of unannotated temporal data to enhance temporal consistency in our detected loops.

\paragraph{A topological score function} We have derived a topological score function from highly interpretable constituents for both loops and components, which split into subscores analogous to precision and recall. These are joined into F1-like scores, again for loops and components individually, before merging them to a final topological score. 

While from a biological point of view, we are primarily interested in the loops, the topological consistency of components is likely to be highly related to the topological consistency of loops, and we thus find it valuable to include both for model selection in order to base our selections on more data. This compositionality of the score makes it highly versatile and interpretable.

\paragraph{Topological model selection} We seek to select those model hyperparameters that produce the best segmentations from a topological point of view. The topological score provides one way to quantify which segmentations are ``best", but alternative scores may produce different ``best" models. Indeed, in the context of hyperparameter tuning (Section~\ref{sec:tuning}), we see, comparing the performance measures in Table \ref{tab:res}, that a given model typically performs best on the particular metric on which its hyperparameter has been tuned. This emphasizes the importance of selecting a topological score function that fits the application well.

\paragraph{Topological loss functions} In our paper, the enforcement of topological consistency from the network's point of view was restricted to tuning the thresholding hyperparameter, while the neural network training used a standard voxel-wise cross-entropy loss. This differs from recent works deriving topological loss functions that encourage topological consistency also in the training of the segmentation network (\cite{shit2021cldice,hu:neurips:2019,clough2020topological,hu2020topology}). We consider having a topological-preserving loss to be a very promising strategy, as it has been demonstrated to work well on a number of simpler segmentation problems. In our experiments, however, we find that topology-preserving loss functions were not sufficiently powerful: the U-net trained with the pixel-wise loss and hyper-parameters tuned on our topology score outperforms the U-net trained using the clDice loss across all measures (see Table \ref{tab:res} and Figure \ref{fig:example_res}). 

We believe that this is caused in part by the challenging nature of our segmentation problem. The signal-to-noise ratio in our dataset is low and  we have a number of artifacts. Moreover, our tubular structures appear at very different scales, meaning that training with small patches, which is e.g.~performed by ~\cite{hu:neurips:2019} to obtain scalability, would likely have trouble capturing all scales. These issues are also likely the reason behind the suboptimal results obtained using the soft-skeletonization algorithm of \cite{shit2021cldice} on our data (see Figure \ref{fig:example_res}).

Note that both the signal-to-noise-ratio and the scale challenge are different from what we typically see in e.g.~vessel- or road segmentation. We believe this is also part of the explanation why methods developed for such tasks are challenged on our dataset.

\paragraph{Segmentation models and data}
Out of the three architectures we experimented with, the semisupervised U-net has proven to outperform the other two, regardless of which performance measure is used (see Table \ref{tab:res}). The joint optimization of the autoencoder and the U-net seems to have driven latent space representations to be more suitable for the segmentation task, compared to the pretrained AE+U-net which acts as an initialization for the U-net. Slightly more surprising is the fact that AE+U-net is outperformed by a randomly initialized U-net. However, this was observed over multiple training runs, indicating that pretraining actually led to a worse initialization. Similar effects have also been observed for natural images in \cite{he2019rethinking}. We speculate that this is due to the artifacts present in our data: The autoencoder may emphasize reconstruction of artifacts, which does not behave as conventional noise.

The convolutions in all our architectures were in 2D. While the 3D U-net of \cite{cciccek20163d} utilizes useful spatial structure, it also has more parameters, which with our limited labeled data led to poorer performance than the 2D counterpart.

Our annotation strategy assumed that a few, detailed annotations of full 3D images would be sufficient to train a good segmentation model, and we chose to focus on having more variance in the validation and test images. To verify that this was a sound assumption, we experimented with swapping the role of the validation and training sets, i.e.~training the 2D U-net on the validation set, and using the training set to tune its hyperparameters. The results were again evaluated on the test set. However, the performance did not improve compared to the U-net reported in Table~\ref{tab:res}.

Although the semisupervised U-net utilized the diversity of unlabeled data to improve segmentation, its predictions are not perfect. As shown in Figure~\ref{fig:difficulty}, our performance correlates negatively with segmentation entropy as well as with segmentation difficulty as assessed by the trained expert. Thus, endowing the extracted tubular network with a notion of uncertainty could ensure safe interpretation.

\paragraph{Loop tracking}
Using temporal information to filter out loops which were present only a short time, helped increase loop score drastically (see Sec.~\ref{sec:loop}). This postprocessing step has proved useful in coping with artifacts and noise in our data.

As mentioned before, having accurate detection and tracking of tubular loops is biologically important. Thus, we evaluated our model performance against expert opinion. Table \ref{tab:hota} provides us with both detection and tracking performances. One can observe that the detection precision is higher than detection recall, meaning there are comparably more FN than FP. This is also partially supported by the results in Fig.~\ref{fig:pia}.

\paragraph{Conclusion and Outlook}
In this paper, we tackled segmenting pancreatic tubular networks, which are dissimilar to other anatomical networks in several aspects. In doing so, we proposed a novel and highly interpretable topological score function, which we used for evaluation as well as model selection. We also compared various training schemes, showing that it is advantageous to incorporate unlabeled data to learn more generalizable latent features. Finally, we showed that postprocessing the detected loops by applying multi-object tracking over time to greatly reduced false positive loops.

Compared to the segmentation of common tubular structures such as vessels or neurons, our images are challenging and our segmentations are imperfect. Recall, however, that the purpose of our segmentation is to provide topological features for testing hypotheses regarding the development of pancreatic organs. One alternative to automatic segmentation is manual segmentation of individual time point volumes, which is too time consuming to be feasible. The other remaining alternative is to manually annotate the topological features (existence of loops and beta cells). This forms the basis of previous biological papers such as ~\cite{kesavan2009cdc42} and ~\cite{bankaitis2015feedback}. However, such manual annotation, which is motivated by finding a dependency between tubular structure and nearby beta cells, is also at risk of being biased: A human who expects to find cells near loops, may also have a tendency to over-annotate loops near cells. An automatic method for quantifying loops does not suffer from this bias. Therefore, an imperfect segmentation still may have great value in its ability to confirm -- or not confirm -- the underlying biological hypothesis in an unbiased way.

Our segmentation problem is challenging, and in addressing it, we have experienced that simple, established methods are more robust than complex ones. By remaining in the realm of established methods, we are able to define a topological score which is also faithful to geometry, which we believe adds robustness. Going forwards, we hope that utilizing geometry to strengthen the matching of topological features, may form a starting point for more robust loss functions that can also be used to train topologically consistent models.

%%%%%%%%%%%%%%%%%%%%%%%%%%%%%%%%%%%%%%%%%%%%%%%%%%%%%%%%%%%%%%%%%%%%%%%
% Mandatory Sections. Please complete, especially for final publication
%%%%%%%%%%%%%%%%%%%%%%%%%%%%%%%%%%%%%%%%%%%%%%%%%%%%%%%%%%%%%%%%%%%%%%%

% Acknowledgements.
% Please include any funding, intellectual contributions not included in the authorship, and any other acknowledgements.
\acks{This work was supported by the Novo Nordisk Foundation through project grant NNF17OC0028360 as well as through the Center for Basic Machine Learning Research in Life Science (NNF20OC0062606). The work is also supported by the Pioneer Centre for AI, DNRF grant number P1.}

% Ethical Standards.
% Please edit with the appropriate ethics considerations for your work. Include any pertinent IRB information, etc.
%
% Please note that the submission requirements included:
% The work presented must follow appropriate ethical standards in conducting research and writing the manuscript, following all applicable laws and regulations regarding treatment of animals or human subjects.
\ethics{The work follows appropriate ethical standards in conducting research and writing the manuscript, following all applicable laws and regulations regarding treatment of animals or human subjects.}

% Conflict of Interest
% Declaration of possible conflicts of interest: Authors must disclose any financial, organisational, commercial or personal conflicts of interest that might bias their work.
% If no conflicts, please say "We declare we don't have conflicts of interest."
\coi{We declare we don't have conflicts of interest.}

\bibliography{melba}

\end{document}